\RequirePackage{fix-cm}
\documentclass[twocolumn]{svjour3}          
\smartqed  

\usepackage{graphicx,amsmath,amsfonts,amssymb,amsthm,epsfig,epstopdf,url,array,xcolor,soul,subfigure,filecontents}
\usepackage{textcomp}
\usepackage[misc]{ifsym}
\usepackage[numbers]{natbib}
\usepackage{tabularx, booktabs, makecell, caption}
\usepackage{siunitx}
\begin{document}

\title{Multi-Robot Coordination and Planning in Uncertain and Adversarial Environments
}


\author{Lifeng Zhou         \and
        Pratap Tokekar 
}


\institute{Lifeng Zhou  \at
         Electrical \& Computer Engineering, Virginia Tech\\
              Blacksburg, VA 24061, USA \\
              \email{\texttt{lfzhou@vt.edu; lfzhou@seas.upenn.edu}}           
           \and
           \Letter ~ Pratap Tokekar \at
        Computer Science, University of Maryland \\ College Park, MD 20742, USA \\
    \email{\texttt{tokekar@umd.edu}}
}

\date{Received: date / Accepted: date}

\maketitle

\noindent\textbf{Abstract}

\noindent\textbf{Purpose of review} Deploying a team of robots that can carefully coordinate their actions can make the entire system robust to individual failures. In this report, we review recent algorithmic development in making multi-robot systems robust to environmental uncertainties, failures, and adversarial attacks.  

\noindent\textbf{Recent findings} We find the following three trends in the recent research in the area of multi-robot coordination: (1) resilient coordination to either withstand failures and/or attack or recover from failures/attacks; (2) risk-aware coordination to manage the trade-off risk and reward, where the risk stems due to environmental uncertainty; (3) Graph Neural Networks based coordination to learn decentralized multi-robot coordination policies. These algorithms have been applied to tasks such as formation control, task assignment and scheduling, search and planning, and informative data collection.

\noindent\textbf{Summary} In order for multi-robot systems to become practical, we need coordination algorithms that can scale to large teams of robots dealing with dynamically changing, failure-prone, contested, and uncertain environments. There has been significant recent research on multi-robot coordination that has contributed resilient and risk-aware algorithms to deal with these issues and reduce the gap between theory and practice. Learning-based approaches have been seen to be promising, especially since they can learn who, when, and how to communicate for effective coordination. However, these algorithms have also been shown to be vulnerable to adversarial attacks, and as such developing learning-based coordination strategies that are resilient to such attacks and robust to uncertainties is an important open area of research.
 
\keywords{Multi-robot systems \and Resilient and risk-aware coordination \and Graph neural networks}

\section{Introduction}
\label{sec:intro}
As the famous adage goes ``there is safety in numbers.'' The objective of this paper is to give an overview of the recent research that is aimed at making teams of robots safer and robust through careful coordination. The last decade has seen multi-robot systems being increasingly used in areas such as manufacturing, warehouse management, agriculture, and environmental monitoring~\cite{roadmap251:online,robothelp}. Deploying a team of robots, instead of a single one, increases the spatial and temporal scales at which the robots are deployed. However, beyond improving the scalability, a team of robots also improve the robustness of the entire system through redundancy and heterogeneity. Through careful coordination between the actions of the robots, we may be able to make the team as a whole robust to uncertainty present in the real-world and to individual failures~\cite{tzoumas2018resilient,zhou2018resilient}. This is a crucial capability as we are starting to have robots coexist with humans and therefore, we need algorithms that can be trusted under real-world, not just ideal conditions. In this paper, we report recent progress in multi-robot coordination and planning that is particularly focusing on increasing the resilience and robustness of the system.

Designing such algorithms, especially to enable coordination between multiple robots under uncertainty and robot failures is a challenging problem. Due to noisy sensing \cite{martinelli2005multi}, imperfect motion, unknown environmental conditions \cite{prorok2019redundant,chung2019risk}, or robot failures \cite{park2018robust}, the robots are very likely to encounter unforeseen events in practice. For example, the travel time for the robots may depend on environmental factors such as traffic and may not be exactly known. As a result, any algorithm that coordinates the actions of the robots will have to operate on uncertain information. Coordination in the presence of uncertainty can be risky \cite{prorok2019redundant,chung2019risk,mataric2003multi}. There is a risk of the actual performance of the robot team during execution time being significantly different from the expected performance during planning time. 

In addition to uncertainty, it is also challenging to deal with catastrophic robot failures either due to natural causes or due to adversarial attacks~\cite{denning2009spotlight,agmon2011multi,sless2014multi}. For example, an adversary can attack the system by spoofing fake identities~\cite{gil2017guaranteeing} or sharing incorrect information~\cite{saulnier2017resilient}. An adversary may also choose specific robots to attack so that the team encounters a worst-case loss in the performance~\cite{schlotfeldt2018resilient}. This is especially crucial in applications such as surveillance and security where counter-Unmanned Aerial Vehicles strategies are being developed \cite{cuas}. In many cases, even if a part of the system fails, the entire system performance can be significantly undermined \cite{tzoumas2018resilient,zhou2018resilient}. 

Pioneering work in this area was done by Parker~\cite{parker1994heterogeneous} who proposed the ALLIANCE framework for coordination in heterogeneous robot teams where individual robots may fail. In this report, we discuss the recent trends, building on this pioneering work, in designing effective multi-robot coordination approaches to address the challenges mentioned above.  Specifically, we focus on three recent trends: 
\begin{enumerate}
    \item \textbf{Resilient coordination} algorithms have been designed to counter adversarial attacks and failures (Section \ref{sec:resilient_coord}). These algorithms can be roughly categorized into two broad approaches. One, researchers focused on designing \textit{robust coordination} approaches that \emph{prepare} systems to withstand attacks \cite{zhou2018resilient,saldana2017resilient} (Section \ref{subsec:resi_control} and \ref{subsec:resi_submodular}). Two, \textit{adaptive and reactive} algorithms have been proposed to enable systems to \emph{recover} from attacks and failures \cite{ramachandran2019resilience,song2020care} (Section \ref{subsec:resilient_configuration}). Resilient coordination algorithms have shown the effectiveness in many robotics tasks, such as team formation \cite{saldana2017resilient}, target estimation \cite{mitra2019resilient}, and data collection \cite{zhou2018resilient,schlotfeldt2018resilient}.
    
    \item \textbf{Risk-aware coordination} aims at addressing the risk of performance loss from uncertainty when a team of robots are operating in uncertain environments (Section~\ref{sec:risk_cord}). One line of recent work is focused on defining and using appropriate \textit{risk measures} of the stochastic team objective (induced by the uncertainty). Several risk-aware coordination algorithms have been developed that allow robot teams to trade-off the reward collected with the risk due to uncertainty \cite{park2018robust,yang2017algorithm,zhou2018approximation} (Section~\ref{subsec:risk_control_opt} and \ref{subsec:risk_submodular}). Another line of work is focused on designing \textit{risk/uncertainty-aware search} algorithms for multi-robot planning with uncertainty \cite{oliehoek2016concise} (Section \ref{subsec:risk_graph_search}). These risk-aware coordination algorithms have been shown to be useful in many robotics tasks, such as team formation \cite{park2018robust,zhu2019chance}, task allocation \cite{yang2017algorithm}, data collection \cite{zhou2018approximation}, and graph search in partially known environments \cite{oliehoek2016concise}. 
    
    \item \textbf{Coordination using Graph neural networks (GNNs)} is a relatively new research direction in multi-robot coordination. By leveraging useful inherent properties of GNNs such as decentralized information propagation, permutation equivalence, and stability \cite{ruiz2020graph}, researchers have shown how to learn \textit{decentralized and optimal} strategies for multi-robot tasks such as team formation \cite{tolstaya2020learning}, path planning \cite{li2019graph}, and task scheduling and assignment \cite{wang2020learning} (Section~\ref{sec:coord_gnn}). One significant advantage of GNN-based approaches is that the strategies learned from smaller cases can be \textit{generalized} to larger settings. For example, these strategies show promise in generalizing across the number of robots~\cite{chen2020multi,prorok2018graph} which can help make the team robust to individual failures. These advantages enable GNN based methods to provide near-optimal solutions for large-scale coordination problems, which might be otherwise  intractable by using classical coordination approaches. 
\end{enumerate}
Based on these research trends, we also discuss several potential research directions for the near term. These include security in deep-learning based coordination methods, risk-aware coordination with the trade-off between local and global interest, and parsimonious communications in GNN based coordination approaches (described further in Section~\ref{sec:future}). 

\section{Resilient Coordination} \label{sec:resilient_coord}
Resilience to unexpected events is a critical capability for getting robots from controlled factory settings to the real world~ \cite{zhang2017resilient}. Resilience broadly refers to two capabilities: (1) preparing a system to be robust enough to withstand faults and attacks; and (2) adapting and recovering from individual failures and attacks. There has been significant work focusing on making individual robots resilient~\cite{bezzo2014attack,bezzo2016stochastic}. Recently, there is a trend on investigating resilience in multi-robot teams, often grounded in tasks such as formation control \cite{saulnier2017resilient,saldana2017resilient,leblanc2013resilient,renganathan2017spoof,saldana2018triangular,guerrero2018design,saldana2019resilient,guerrero2019realization,usevitch2019resilient,usevitch2020resilient,senejohnny2018resilience,senejohnny2019resilience}, wireless communication \cite{gil2017guaranteeing,sun2019resilient}, state estimation \cite{mitra2019resilient,mitra2016secure,mitra2018impact,mitra2019byzantine}, data collection \cite{zhou2018resilient,schlotfeldt2018resilient,zhou2019approximation,zhou2019distributed,shi2020robust}, attack-defense games \cite{shishika2018local,shishika2019team,shishika2020cooperative}, and adaptive reconfiguration \cite{ramachandran2019resilience,song2020care,ramachandran2020ICRA,ramachandran2020}. A summary of these applications is found in Table~\ref{tab:resilient_application}. 

A general goal in resilient multi-robot systems is to guarantee a desired team performance even though some robots in the team fail \cite{song2020care}, are malicious \cite{saulnier2017resilient}, or get attacked \cite{zhou2018resilient}. We summarize the sources of robot failures and attacks in Table~\ref{tab:resilient_mal_source}. To address these challenges, researchers have been focusing on designing various coordination approaches to make multi-robot teams resilient. We describe the main approaches in the following.   

\begin{table*}
\caption{Resilient multi-robot coordination: tasks}
\label{tab:resilient_application}  
\centering
\begin{tabular}{ll}
\hline\noalign{\smallskip}
Formation Control & \cite{saulnier2017resilient,saldana2017resilient,renganathan2017spoof,saldana2018triangular,guerrero2018design,saldana2019resilient,guerrero2019realization,usevitch2019resilient,usevitch2020resilient,senejohnny2018resilience,senejohnny2019resilience}\\
Task Allocation & \cite{mayya2020resilient}\\
Target Tracking & \cite{mitra2016secure,mitra2018impact,mitra2019resilient,mitra2019byzantine,zhou2018resilient,schlotfeldt2018resilient,zhou2019distributed,ramachandran2020ICRA,ramachandran2020resilient}\\
Coverage/exploration & \cite{schlotfeldt2018resilient,ramachandran2020,shi2020robust}\\
Communication & \cite{ramachandran2020ICRA,ramachandran2020resilient,gil2017guaranteeing,sun2019resilient,ramachandran2019resilience}\\
Perimeter defence & \cite{shishika2018local,shishika2019team,shishika2020cooperative}\\
\noalign{\smallskip}\hline
\end{tabular}
\end{table*}

\begin{table*}
\caption{Resilient multi-robot coordination: malfunction sources}
\label{tab:resilient_mal_source}  
\centering
\begin{tabular}{ll}
\hline\noalign{\smallskip}
Spoofing Attack & \cite{gil2017guaranteeing,saulnier2017resilient,saldana2017resilient,mitra2019resilient,renganathan2017spoof,saldana2018triangular,guerrero2018design,saldana2019resilient,guerrero2019realization,usevitch2019resilient,usevitch2020resilient,senejohnny2018resilience,senejohnny2019resilience,mitra2016secure,mitra2018impact,mitra2019byzantine}\\
Robot/Resource Failure & 
\cite{ramachandran2019resilience,sun2019resilient,ramachandran2020ICRA,ramachandran2020resilient}\\
Denial of Service Attack &
\cite{zhou2018resilient,schlotfeldt2018resilient,zhou2019distributed,shi2020robust} \\
Environmental Change & 
\cite{mayya2020resilient}\\
\noalign{\smallskip}\hline
\end{tabular}
\end{table*}

\subsection{Resilient Formation Control and Estimation}~\label{subsec:resi_control}
Formation control is a classical problem for multi-robot systems. The goal is for the robots to achieve a desired formation, maintain it as the formation moves, and reconfigures to a new formation~\cite{oh2015survey}. The standard assumption in classical techniques is that all robots are cooperative. Saulnier et al. presented a \textbf{resilient formation control} approach \cite{saulnier2017resilient} where they designed a control policy that enables a team of mobile robots to achieve desired formations even though some team members are non-cooperative (or adversarial) in the sense that they broadcast malicious/spoofing signals. The authors built on the previous work about the $r$-robustness of communication topology \cite{leblanc2013resilient} and proposed an efficient approach to manage the algebraic connectivity of the graph to guarantee resilient formation. Following this line, resilient formations have been investigated in time-varying communication graphs \cite{saldana2017resilient}, triangular robust graphs \cite{saldana2018triangular}, and triangular and square lattices \cite{guerrero2018design,saldana2019resilient}. Then Usevitch and Panagou studied a resilient leader-follower formation problem in \cite{usevitch2019resilient} where they proposed a resilient controller that allows well-behaving robots to track a reference state even though a bounded subset of leaders and followers are adversarial. In these  studies, the robots update their motion synchronously and periodically. As a contrast, some recent work has focused on the problem of resilient formation if the robots have their
own clocks and may update in an asynchronous way, which makes progress towards designing resilient event-triggered
and self-triggered coordination strategies \cite{senejohnny2018resilience,senejohnny2019resilience}. 

In addition to resilient formations,  researchers have investigated \textbf{resilient state estimation} of a static or dynamical process (target of interest) by a team of robots \cite{mitra2019resilient,mitra2016secure,mitra2018impact,mitra2019byzantine}. In particular, Mitra et al. presented an important approach \cite{mitra2019resilient} where robots are tasked to estimate the state of a dynamical process given the challenges of intermittent measurements, communication loss, and adversarial team members. To cope with these challenges, they designed resilient, distributed algorithms to guarantee correct state estimation in the dynamically changing and adversarial environments. 

\subsection{Resilient Submodular Maximization}~\label{subsec:resi_submodular}

State estimation, described in the previous subsection, belongs to a broader class of information gathering problems. These problems include process or target tracking \cite{tokekar2014multi,atanasov2014information,zhou2017active,zhou2018active,zhou2019sensor}, environmental monitoring \cite{michini2014robotic}, and search-and-rescue \cite{kumar2012opportunities}. In multi-robot information gathering problems, the objective is to maximize the joint information gathered by the robot team, such as the joint area covered or explored. Given that the robots may have overlaps over the collected data, this objective generally turns out to be a \textit{submodular function} \cite{nemhauser1978analysis,fisher1978analysis}. 

\begin{figure}
\centering
 \includegraphics[width=0.33\textwidth]{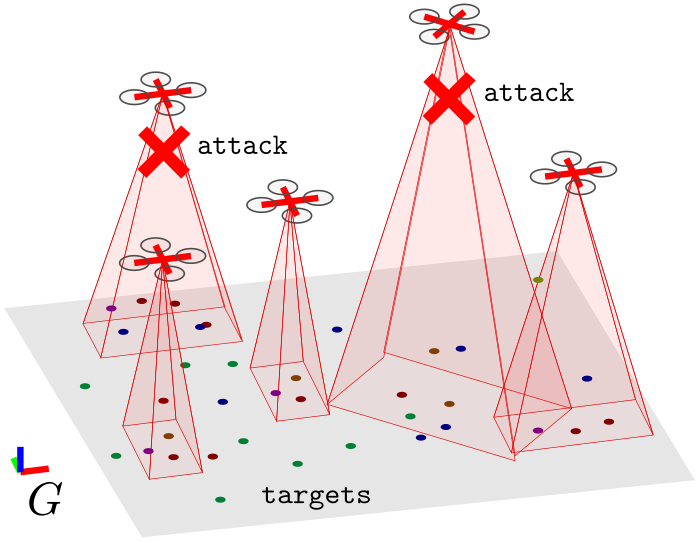}
\caption{Multi-robot target tracking with adversarial attacks. An adversary can block robots' tracking cameras to disable their tracking abilities. (\textcopyright~[2020] IEEE. Reprinted, with permission, from  \cite{zhou2019distributed}).
}
\label{fig:multi-track}       
\end{figure}

Submodularity captures the property of
diminishing returns of set fucntions. Even though maximizing submodular functions is NP-hard, a simple greedy algorithm yields a constant-factor approximation of the optimal solution \cite{nemhauser1978analysis,fisher1978analysis}. However, if the robots operate in an adversarial environment and some of them or their sensors are attacked, the simple greedy algorithm may perform arbitrarily bad \cite{zhou2018resilient}. To this end, Tzoumas et al. presented the first scalable resilient algorithm to counter any number of adversarial (specifically, denial-of-service) attacks or failures \cite{tzoumas2017resilient}. Building on this work, recent studies have investigated designing \textbf{resilient information collection} algorithms in target tracking \cite{zhou2018resilient} and information gathering scenarios \cite{schlotfeldt2018resilient}. Particularly, the resilient target tracking algorithm proposed by Zhou et al. \cite{zhou2018resilient} guarantees a provably close-to-optimal team performance even though some robots in the team are attacked and their tracking cameras are blocked (Figure~\ref{fig:multi-track}). Then the same set of authors extended their resilient target tracking algorithm to a constrained communication scenario where robots have limited communication range and can only communicate locally \cite{zhou2019distributed}. They proposed a distributed robust algorithm where robots first form local groups and then the groups operate the resilient target tracking algorithm \cite{zhou2018resilient} in parallel to counter attacks. Later, Shi et al. applied the resilient algorithm \cite{tzoumas2017resilient} into a resilient team orienteering problem where a group of robots plans paths within a limited budget to collect data in an adversarial environment \cite{shi2020robust}. The authors designed a robust multi-path planning algorithm that allows robots to plan trajectories over a longer time horizon and guarantees provable team data-collection performance, despite a number of robot being attacked or failing.  

\subsection{Resilient Reconfiguration via Optimization}~\label{subsec:resilient_configuration}

In addition to designing coordination algorithms that withstand attacks or failures \cite{zhou2018resilient,zhou2019distributed,shi2020robust,tzoumas2017resilient}, we also need \textbf{resilient reconfiguration} approach that enables robot teams to adaptively recover after attacks or faults \cite{ramachandran2019resilience,song2020care,ramachandran2020ICRA,ramachandran2020}. Ramachandran et al. studied the problem of maintaining resource availability in a network of multiple robots \cite{ramachandran2019resilience} in such conditions. The resources can be sensing or computational capabilities provided by the robots. The authors designed a resilient resource reconfiguration framework that enables robots to maintain access to the resources by effectively reconfiguring inter-robot communication networks if their resources are not available due to failures. Then this resilient reconfiguration framework was utilized for maintaining sensing quality for robots to track targets \cite{ramachandran2020ICRA}. Later, a resilient multi-robot coverage framework was designed in \cite{ramachandran2020} where well-functioning robots adaptively reposition themselves to maintain a good team coverage performance once a robot in the team fails. Further, to completely cover or explore an environment by a team of robots, Song et al. presented a distributed event-driven replanning algorithm to adaptively assign tasks to compensate for the team loss induced by robot failures \cite{song2012simultaneous}. Specifically, a game-theoretic structure was designed to trigger resilient task reallocations for the well-behaving robots, e.g., either keeping performing their own tasks or helping the failed robot to perform its task.  

\section{Risk-Aware Coordination} \label{sec:risk_cord}
The resilient coordination approaches discussed heretofore seek to optimize for the worst-case performance. While in most cases, the focus is on being resilient to adversarial attacks, these approaches can also be used to find a conservative plan that is resilient to uncertainty that is present in most practical settings. However, optimizing for the worst-case may be too conservative in such settings. Instead, one may want approaches that can trade-off reward versus risk due to uncertainty. The uncertainty can stem from imperfect information about internal information about the robots, noisy sensing, imperfect motion \cite{martinelli2005multi} as well as from external sources such as unknown or partially known environments \cite{zhou2018approximation,chung2019risk}. A summary of various sources of uncertainty is given in Table \ref{tab:riskaware_uncer_source}. Because of the uncertain information, the actual robot system's performance in execution can significantly diverge from the expected performance at the planning stage, which puts the system's performance at risk. 

\begin{figure}
  \centering
  \includegraphics[width=0.8\columnwidth]{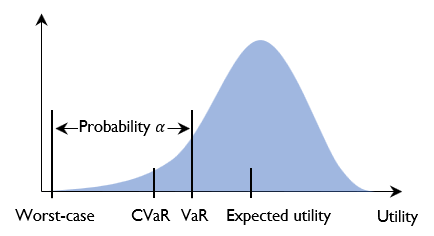}
  \caption{An illustration of several risk measures. }
  \label{fig:riskmea}
\end{figure}

The standard approach of dealing with uncertainty is to consider either the expected-case performance~\cite{prorok2019redundant,peltzer2020stt} or the worst-case performance~\cite{zhou2017active,zhou2018active,yel2018self}. These represent extremes; recently, there is a trend of using more nuanced measures that better capture risk. In particular, these include mean-variance \cite{chung2019risk,toubeh2019risk}, chance-constraints or Value-at-Risk (VaR) \cite{yang2017algorithm}, and Conditional-Value-at-Risk (CVaR) \cite{chow2015risk,chow2017risk,zhou2018approximation} (Figure~\ref{fig:riskmea}). Specifically, the mean-variance measure has the form $\mathbb{E}(U) + \lambda\sigma^2(U)$ where $\mathbb{E}(U)$ and $\sigma^2(U)$ are the expectation and the variance of the stochastic performance (or utility) $U$, and $\lambda$ is a weighting parameter deciding the relative importance we place on the expected performance and the risk (in this case, the variance). The Value-at-Risk $\text{VaR}_\alpha (U)$ denotes the $\alpha$-quantile of stochastic utility $U$ and is defined as:
$$
\text{VaR}_{\alpha}(U) = \text{inf} \{\tau\in\mathbb{R}, ~\text{Pr} [U\leq \tau] \geq \alpha\},
$$  
with $\alpha \in [0,1]$ denoting the user-prescribed risk tolerance level.
Based on \text{VaR}, the Conditional-Value-at-Risk $\text{CVaR}_{\alpha}(U)$ is defined as: 
$$
\text{CVaR}_{\alpha}(U) = \underset{y}{\mathbb{E}}[U|U\leq \text{VaR}_{\alpha}(U)],
$$
where $y$ is a random parameter representing the uncertainty. Thus, CVaR measures the expected performance in the $\alpha$ worst fraction of cases. A summary of commonly used risk measures with a comparison in terms of six proposed axioms has been described in \cite{majumdar2020should}. 

Based on these risk measures, researchers have developed risk-aware approaches in various robotics tasks such as graph search and motion planning \cite{chow2015risk,chow2017risk,chung2019risk,fridovich2020confidence}, controls \cite{park2018robust,singh2018framework,zhu2019chance}, task allocation and assignment \cite{prorok2019redundant,yang2017algorithm,yang2018algorithm,yang2020chance}, information collection \cite{jorgensen2018team,zhou2018approximation}, and machine learning \cite{chow2017risk,lacotte2019risk}. Here, we focus on the risk-aware approaches for multi-robot systems. We describe these approaches in the following and summarize the corresponding applications in Table~\ref{tab:riskaware_application}.

\begin{table*}
\caption{Risk-aware multi-robot coordination: uncertainty sources}
\label{tab:riskaware_uncer_source}  
\centering
\begin{tabular}{llllllll}
\hline\noalign{\smallskip}
Random Robot/Sensor Failure  & \cite{park2018robust,zhou2018approximation} \\
Uncertain Travel Time/Distance & \cite{nam2016analyzing,zhou2018approximation,yang2018algorithm,peltzer2020stt}    \\
Dynamic or Unknown Obstacles & \cite{zhu2019chance,zhu2019b,da2019collision,wang2020non,yang2020chance,fridovich2020confidence,indelman2018cooperative}   \\
Generic Partial Observability  & \cite{kochenderfer2015decision,chow2015risk,oliehoek2016concise,amato2016policy,omidshafiei2017decentralized,amato2019modeling,omidshafiei2017deep,chow2017risk}  \\
Extraction Uncertainty by DL & \cite{Toubeh2019risk2,sharma2020risk}\\

\noalign{\smallskip}\hline
\end{tabular}
\end{table*}

\begin{table*}
\caption{Risk-aware multi-robot coordination: tasks and techniques}
\label{tab:riskaware_application}  
\centering
\begin{tabular}{llllllll}
\hline\noalign{\smallskip}
Formation control & \cite{park2018robust} \\
Task assignment & \cite{nam2016analyzing,yang2017algorithm,yang2018algorithm,yang2020chance,sharma2020risk} \\
Coverage/exploration & \cite{zhou2018approximation,jorgensen2018team,chung2019risk} \\
Collision avoidance & \cite{zhu2019chance,zhu2019b,da2019collision,wang2020non}\\
Graph search & \cite{yang2020chance,jorgensen2018team,chung2019risk,fridovich2020confidence,kochenderfer2015decision,chow2015risk,oliehoek2016concise,amato2016policy,omidshafiei2017decentralized,amato2019modeling,indelman2018cooperative,peltzer2020stt,Toubeh2019risk2,sharma2020risk}\\
Learning & \cite{Toubeh2019risk2,sharma2020risk,omidshafiei2017deep,chow2017risk,lacotte2019risk}\\
\noalign{\smallskip}\hline
\end{tabular}
\end{table*}

\subsection{Risk-Aware Control and Optimization} \label{subsec:risk_control_opt}

Researchers have recently investigated risk-awareness in multi-robot control and optimization such as formation control \cite{park2018robust}, chance-constrained optimization \cite{zhu2019chance,yang2017algorithm}, and CVaR based  optimization \cite{nam2016analyzing}. Particularly, Park and Hutchinson designed distributed robust controllers to guarantee the rendezvous for multi-robot systems even though some robots in the team randomly fail \cite{park2018robust}. They considered both the mean-variance and the worst-case measures for the stochastic cost function. Zhu and Alonso-Mora used a chance-constraint measure, explicitly constraining the probability of an undesirable event, to guarantee the safe navigation (in a probabilistic sense) of micro-air vehicles in cluttered and dynamically changing environments \cite{zhu2019chance}. They formulated a chance-constrained nonlinear model predictive control problem that takes the probability of collision as constraints. For this problem, they proposed three coordination strategies based on different communication settings and evaluated the effectiveness of proposed methods through real-world experiments. Later, the authors utilized buffered uncertainty aware Voronoi cells, computed by satisfying a set of chance constraints, to guarantee the inter-robot collision avoidance in a probabilistic way \cite{zhu2019b}. The chance constraints have also been utilized in non-convex collision-free path planning \cite{da2019collision} and for stochastic planning problems that go beyond the classical Gaussian uncertainty \cite{wang2020non}.  

Risk-awareness has also been considered in task assignment problems. Yang and Chakraborty studied a chance-constrained combinatorial optimization problem that takes into account the risk in multi-robot assignment \cite{yang2017algorithm}. They later extended the chance-constrained formulation to knapsack problems \cite{yang2018algorithm}. They solved the problem by transforming it into a risk-aware problem with mean-variance measure \cite{park2018robust}. Instead of chance-constrained measure, Nam and Shell analyzed the sensitivity of assignment optima by using both the expectation and CVaR measures in a multi-robot task allocation problem \cite{nam2016analyzing}.    

\subsection{Risk-Aware Submodular Maximization}~\label{subsec:risk_submodular}
As described in Section~\ref{subsec:resi_submodular}, many multi-robot coordination objectives are naturally submodular. Existing work on submodular optimization has focused on the deterministic case; however, in practice the objective can be stochastic (e.g., random robot or sensor failures \cite{park2018robust,zhou2018approximation} or uncertain travel time because of unknown traffic \cite{zhou2018approximation,prorok2019redundant}). 

Using expectation as the measure, Prorok studied the problem of assigning multiple robots to goal locations under travel-time uncertainty \cite{prorok2019redundant}. This approach assigned redundant robots to goal locations to counter uncertainty, which makes the objective, the total waiting time at goal locations, a supermodular function ($f(\mathcal{S})$ is supermodular if $-f(\mathcal{S})$ is submodular). Since the expectation of a supermodular function is still supermodular, a simple greedy algorithm was utilized to efficiently minimize the total waiting time. The expectation was also used to measure the stochastic submodular team performance in a team surviving orienteering problem where a team of robots is tasked to traverse a dangerous environment with a probability of survival for each robot \cite{jorgensen2018team}. The authors proposed an approximate greedy approach with a provable guarantee for selecting risk-aware paths for the robots. They also investigated the problem in an online version and with heterogeneous teams and verified the proposed algorithm in large-scale real-world scenarios. 

While optimizing the expected performance has its uses, it also has its pitfalls since the expectation measure is risk-neutral and may not work well in some extreme (bad) cases \cite{zhou2018approximation,majumdar2020should}. At the same time, decisions made considering worst-case scenarios can be conservative~\cite{zhou2017active,zhou2018active,zhou2018resilient}. Recent work optimize for risk measures such as CVaR and aim to fill this gap between expected-case and worst-case analysis~\cite{rockafellar2000optimization,majumdar2020should}. By optimizing CVaR with a user-defined risk parameter $\alpha$, the robot team can balance the trade-off between the reward and the risk it would like to take \cite{zhou2018approximation}. 

However, when the objective is a discrete (stochastic) submodular function,
Maehara presented a negative result for optimizing $\text{CVaR}$ \cite{maehara2015risk}--- there is no \\ polynomial-time multiplicative approximation algorithm under some reasonable assumptions in computational complexity. To circumvent this issue, Ohsaka and Yoshida adopted an idea from {portfolio optimization} and proposed a method of selecting a distribution over available sets rather than selecting a single set, and gave a provable guarantee \cite{ohsaka2017portfolio}. Following this line, Wilder considered a CVaR maximization of a {continuous submodular} function instead of the submodular set functions \cite{wilder2018risk}. They gave a $(1 - 1/e)$--approximation algorithm for {continuous submodular} functions and also evaluated the algorithm for discrete submodular functions using {portfolio optimization} \cite{ohsaka2017portfolio}. Instead of selecting a   portfolio over the available sets, Zhou and Tokekar presented a risk-aware sampling-based algorithm with a bounded approximation to select a single set for CVaR based discrete submodular optimization \cite{zhou2018approximation,zhou2019risk}. They demonstrated the effectiveness of the risk-aware algorithm through two case studies: sensor placement with random failures and vehicle assignment for mobility-on-demand under travel-time uncertainty (with an online version presented in \cite{zhou2019risk}). Their risk-aware algorithm has also been utilized for planning risk-aware paths for the robots with uncertainty extracted from Bayesian deep learning models \cite{Toubeh2019risk2,sharma2020risk} and extend to the case of planning a route~\cite{balasubramanian2020risk}.

\subsection{Uncertainty-Aware Search and Planning} \label{subsec:risk_graph_search}

Classical planning problems have been extensively studied \cite{lavalle2006planning}. Based on the assumption that the robot actions and search space are deterministic, researchers have developed many powerful planning algorithms such as A* \cite{hart1968formal} and RRT* \cite{karaman2011sampling} to find optimal start-to-goal paths. However, after incorporating uncertainty from the robot state and/or the world model, these planning problems become challenging \cite{kochenderfer2015decision,chung2019risk}. Some recent work has investigated incorporating the uncertainty in the classical planners to tackle these problems \cite{hollinger2016learning,toubeh2019risk}. 

A typical way to address the uncertainty in planning is to formulate these problems as Partially Observable Markov Decision Processes (POMDPs) that consider uncertainty in robot states, actions, and observations \cite{monahan1982state,oliehoek2016concise}. Particularly, for multi-robot coordination, Amato et al. presented a MacDec-POMDP planning algorithm to deal with uncertain sensing and limited communication in multi-robot planning \cite{amato2016policy}. They showed several  properties of the proposed MacDec-POMDP planning algorithm and its advantages in solving larger problems over existing Decentralized POMDPs (Dec-POMDP) planners. Later, to solve multi-robot planning problems in continuous spaces, Omidshafiei et al. extended the Dec-POMDP to the Decentralized Partially Observable Semi-Markov Decision Process (Dec-POSMDP) using task macro-actions that allow robots to make decisions asynchronously \cite{omidshafiei2017decentralized}. They also proposed scalable algorithms to generate robust solutions for solving Dec-POSMDPs. If the underlying Dec-POMDP model is not assumed to be known a priori or a full simulator is available at planning time, Liu presented a policy-based reinforcement learning approach that updates policies though agents interacting with the environment \cite{liu2016learning}. They also showed the proposed approach can generate valid macro-action controllers (used in \cite{amato2016policy,omidshafiei2017decentralized,amato2019modeling}) and learn optimal policies.  

Other related work on multi-robot planning under uncertainty includes belief space planning for navigation in unknown environments \cite{indelman2018cooperative}, path planning with travel-time uncertainty \cite{peltzer2020stt}, task allocation with completion uncertainty \cite{choudhury2020dynamic}, and multi-task
reinforcement learning under partial observability \cite{omidshafiei2017deep}. 

\section{Recent Trend: Coordination by Graph Neural Networks} \label{sec:coord_gnn}
There is a growing trend of using learning-based methods, in particular Graph Neural Networks (GNNs) for multi-robot coordination \cite{prorok2018graph,gama2020graph}. These approaches have been shown to learn \emph{what}, \emph{when}, and \emph{who} to communicate with depending on the task at hand, instead of following a rigid communication topology~\cite{liu2020when2com,liu2020who2com}. GNNs  capture the interactions among robots by modeling their coordination as a graph where robots (nodes) share information with their neighbors through communion links (edges). GNN architectures also exhibit the transferability property because of their permutation equivalence and stability, which allows the learned policies to be generalized to previously unseen scenarios \cite{ruiz2020graph,gama2019stability}. Researchers have recently implemented GNNs to learn decentralized and close-to-optimal solutions for the classical problems in multi-robot systems such as formation control \cite{prorok2018graph,tolstaya2020learning,khan2020graph,khan2019graph}, multi-robot path planning \cite{li2019graph}, and task assignment and scheduling \cite{wang2020learning}. 

Particularly, some studies have investigated learning decentralized controllers for large networks of mobile robots to achieve desired formations by either using GNNs to imitate centralized controllers with global information \cite{tolstaya2020learning} or employ GNNs to parameterize policies that are updated by policy gradients \cite{sutton2018reinforcement}. Similarly, applying GNNs that allow robots to communicate with multi-hop neighbors, Li et al. have investigated generating collision-free paths for multiple robots from start positions to goal positions in cluttered environments \cite{li2019graph}. They have shown that, by imitating an expert algorithm, their learned planner that uses only local communication and observations, achieves close-to-optimal performance and is able to generalize to larger robot teams. 

GNN based approach has also been proposed for solving multi-robot task scheduling and assignment that is modeled as a \textit{combinatorial optimization} problem \cite{wang2020learning}. Since combinatorial optimization problems are generally NP-hard, obtaining the optimal solution is computationally intractable for large-scale cases. Even though there exist many approximation algorithms (or heuristic) that runs in polynomial time, they only provide approximation guarantees or solutions without any optimality guarantee. To this end, Wang et al. combined imitation learning (i.e., imitate optimal solvers for small problems) with graph attention networks to learn fast, near-optimal scheduling that is scalable and generalizable \cite{wang2020learning}. 

Learning-based coordination can also lead to interesting emergent behavior in the team of robots~\cite{liu2019emergent,blumenkamp2020emergence,chen2020multi}. Chen et al.~\cite{chen2020multi} showed emergent multi-robot behavior when learning to persistently monitor environments. Specifically, they observed that the learned policies lead to natural partitioning of the environment amongst the robots as well as lead to periodic trajectories that are to be expected in persistent monitoring. Blumenkamp and Prorok found that when a robot is trained to optimize for its own interest, it learns to communicate adversarial information to the other robots (that are optimizing the cooperative interest) \cite{blumenkamp2020emergence} in coverage tasks. This opens up several interesting questions such as whether the agents can \emph{learn} to be resilient to such adversarial information sharing.

\section{Conclusions} ~\label{sec:future}

We outlined several recent work on multi-robot coordination specifically aimed at dealing with adversaries, failures, and environmental uncertainty. While there has been significant developments in making teams of robots resilient and risk-aware, there remain several outstanding issues. In the following, we highlight three directions that we believe are going to be important going forward. 

\noindent\textbf{Secure Intelligent Multi-Robot Systems}~ The state-of-the-art deep learning based methods have shown a great promise in the field of robotics. However, when operating in safety-critical situations, these intelligent learning systems can be easily spoofed or misled (e.g., by uncertainties or adversarial attacks) which can cause unsafe situations \cite{Kurakin2017,eykholt2018robust}. Although some research has focused on designing robust deep neural networks to deal with adversarial perturbations in either the training stage or testing stage \cite{madry2017towards,athalye2018obfuscated,tramer2017ensemble}, there has been little work in the context of multi-robot coordination. As highlighted earlier, recent studies have shown emergence of adversarial communication in multi-robot reinforcement learning~\cite{blumenkamp2020emergence}. As such, it is critical to study how to secure such multi-robot systems to adversarial attacks and guarantee a provably good team performance. Unlike say computer vision tasks, here one can leverage \textit{inter-robot coordination} to mitigate such attacks as has been shown in the recent work. We need to build on this and extend it to the cases where the coordination between agents is learned instead of designed by classical algorithms.

\noindent\textbf{Collaborative Decision Making with Risk Trade-off  
} ~The second future direction is to investigate how individual robots trade-off their local risks with the global/team risk to optimize team performance. For example, in a multi-robot search and rescue scenario, the robot team may want to explore the environment as much as possible while maintaining global connectivity. However, an individual robot may cares about its own safety and energy, e.g., reducing the risk of collisions or traveling through rugged terrains. To do so, it may increase the risk of disrupting team connectivity when the robot selects a safer yet far-away path (from other team members) to bypass obstacles. This can also lead to the risk of undermining team performance by affecting the decisions of other team members. That is because, to maintain global connectivity, other team members will have to abandon their high-reward paths (e.g., the paths can jointly cover a larger area), in order to move closer to this robot. Thus, there is a need for risk-aware algorithms to maximize team performance with the consideration of local and global risk trade-off.

\noindent\textbf{Parsimonious Communications for Multi-Robot Coordination using GNNs} ~As described in Section~\ref{sec:coord_gnn}, GNN based methods have shown promise in learning close-to-optimal solutions by imitating an optimal solver or an expert for multi-robot coordination \cite{tolstaya2020learning,li2019graph,wang2020learning}.
Even though GNN based architecture only requires robots communicating with certain hop neighbors, some of these communications may not be necessary. Some studies have presented parsimonious communication strategies that are selective in \textit{when to communicate with neighbors} \cite{zhou2017active,zhou2018active,liu2020when2com} and \textit{which neighbors to communicate with} \cite{liu2020who2com} to cutoff unnecessary communications among neighbors. Built on this idea, the third future avenue can be embedding parsimonious communication protocols into GNN architectures for multi-robot coordination. This will be particularly important when there is heterogeneity in the sensing, actuation, and computing capabilities of the robots as well as when there are multiple tasks that the robots need to solve for. Not all tasks may require all resources. Learning task-oriented coordination is an important direction of research.

\begin{acknowledgements}
The authors would like to thank the National Science
Foundation (NSF IIS-1637915) and the Office of Naval Research (ONR N00014-18-1-2829) for their supports.
\end{acknowledgements}

\section{Compliance with Ethical Standards}
\noindent\textbf{Conflict of Interest} ~ The authors declare that they have no conflict of
interest. 

\vspace{3mm}

\noindent \textbf{Human and Animal Rights and Informed Consent} ~ This article does not contain any studies with human or animal subjects performed by any of the authors.

%
%

\renewcommand{\bibpreamble}{Papers of particular interest, published recently, have been highlighted as:\\
• Of importance \\
•• Of major importance}
\bibliographystyle{IEEEtran}      
\bibliography{refs}   

\end{document}